\documentclass{article}
\usepackage{spconf,amsmath,graphicx}
\usepackage{multirow}
\usepackage{mathrsfs}
\usepackage{booktabs}

\title{A Glance is Enough: Extract Target Sentence By Looking at A keyword}
\name{Ying Shi$^1$, Dong Wang$^2$, Lantian Li$^3$, Jiqing Han$^1$
\thanks{
  Corresponding authors: D.W. (wangdong99@mails.tsinghua.edu.cn) and J.H. (jqhan@hit.edu.cn)
  }
}
\address{
  $^1$School of Computer Science and Technology, Harbin Institute of Technology, China \\
  $^2$Center for Speech and Language Technologies, BNRist, Tsinghua University, China  \\
  $^3$School of Artificial Intelligence, Beijing University of Posts and Telecommunications, China \\
}
\begin{document}
\ninept
\maketitle
\begin{abstract}
This paper investigates the possibility of extracting a target sentence from multi-talker speech using only a keyword as input. For example, in social security applications, the keyword might be ``help", and the goal is to identify what the person who called for help is articulating while ignoring other speakers. To address this problem, we propose using the Transformer architecture to embed both the keyword and the speech utterance and then rely on the cross-attention mechanism to select the correct content from the concatenated or overlapping speech. Experimental results on Librispeech demonstrate that our proposed method can effectively extract target sentences from very noisy and mixed speech (SNR=-3dB), achieving a phone error rate (PER) of 26\%, compared to the baseline system's PER of 96\%.

\end{abstract}

\section{Introduction}
\label{sec:intro}

Automatic Speech Recognition (ASR) has significantly advanced, even surpassing human performance in some instances~\cite{xiong2016achieving}. However, recognizing speech when two or more speakers are talking concurrently is a challenging task. 
This is referred to as the multi-talker problem and is tackled within two research fields: speaker diarization~\cite{park2022review}, which identifies words spoken by the same speaker from different locations, and overlapping(aka mixed) ASR~\cite{chen2017progressive}, focusing on recognizing each speaker's words during simultaneous speech.
Both fields have witnessed considerable advancements in recent years.

A primary challenge in multi-talker ASR tasks is the necessity to recognize sentences spoken by \emph{all} speakers.
This presents a significant challenge, as even humans find it challenging to listen to two speakers simultaneously~\cite{kawata2020neural}. 
Target-speaker ASR (TS-ASR), a promising solution, enables the system to focus on a particular speaker and only recognize his/her speech. 
This approach is the central idea of TS-ASR~\cite{wang2019voicefilter,vzmolikova2019speakerbeam,moriya2022streaming,zhang2023conformer}. 
In contrast to conventional multi-talker ASR, TS-ASR eliminates the need to know the number of speakers in the signal and avoids the target-permutation problem, which is a significant challenge when training multi-talker models.

TS-ASR utilizes speaker information as the ``attention bias'' and is suitable for monitoring a pre-registered target. 
However, in certain circumstances, the identity of the speaker may not be as important as the content of their speech. 
For example, in public security applications, the word ``help" may be a crucial keyword, and once it is triggered, we need to know what is happening. 
In such situations, at least two aspects are critical: (1) identifying the keyword from possibly mixed and noisy speech, and (2) recognizing the words/sentences of the individual who triggered the keyword whilst ignoring the speech of others. 
This new task, following the nomenclature of TS-ASR, can be termed target-content ASR (TC-ASR).
Both TC-ASR and TS-ASR employ an attention bias, simplifying the challenge compared to full multi-talker ASR.
However, using a keyword as the attention bias is weaker than using speaker characteristics, raising questions about how the keyword guides the decoder to select target content: by semantic continuity or speaker similarity?

This paper aims to investigate the possibility of TC-ASR. 
The overall diagram of the proposed model is depicted in Fig.~\ref{fig:overview}, which comprises a speech encoder and a keyword encoder, both in the form of Transformers. 
The above two encoders are integrated by a cross-attention, and the model is trained using CTC loss.
Surprisingly, our experiments demonstrate that this straightforward model can extract complete sentences by simply considering a single keyword. Further analysis reveals that the model accomplishes this by inferring speaker information, and extracting the sentence spoken by the speaker.
To the best knowledge of the authors, this is the first study that shows a keyword can be used as attention bias and signal a neural model to identify the target speech. 


\section{Related Work}

This work is closely related to overlapping ASR, and the most straightforward approach is to concatenate speech separation (SS) and ASR components sequentially, where multiple ASR decoders are designed to handle multiple speakers.
However, this simple stitch is suboptimal, as the speech produced by the SS module may be distorted. 
This issue can be addressed either by fine-tuning the ASR model using the SS output~\cite{wu2021investigation} or by addressing SS and ASR with a single model trained directly with the ASR loss~\cite{yu2017recognizing}, potentially considering SS as an auxiliary task~\cite{settle2018end}.

No matter in which way, a central problem when training an ASR model with mixed speech is how to deal with speaker permutation, i.e., determining which word is assigned to which decoder branch.
A possible way is to design an assignment rule. 
For instance, Chao et al.~\cite{7122291} assign louder speech to the first branch and weaker speech to the second, and Lu et al.~\cite{lu2021streaming} assign the speaker who first appears to the first branch and the second speaker to the second branch. 
This rule-based assignment has limitations, as in some instances, it may be challenging to determine which speech is louder or which speaker occurs first.
Permutation invariant training (PIT) presented by Yu et al.~\cite{yu2017permutation}, offers a more elegant solution to this problem.
The central concept involves selecting the minimal loss from all potential permutations.
PIT was used firstly for speech separation~\cite{yu2017permutation,kolbaek2017multitalker}, and was quickly applied to train ASR models for overlapping speech~\cite{yu2017recognizing}. 
Multi-branch decoder trained by PIT has become a standard framework for overlapping ASR, and numerous studies have been conducted within this architecture~\cite{qian2018single,meng2023sidecar}.

A particular limitation of the traditional PIT method is its need for prior knowledge about the number of speakers in the overlapping speech signal, and its inability to manage a large number of speakers.
A recent study on serial output training (SOT)~\cite{kanda2020serialized} has overcome this limitation by sequentially outputting the sentences of each speaker, separated by a speaker change (SC) token. 
When integrated with an Attention-based Encoder-Decoder (AED) architecture, SOT enables the model to sequentially output the spoken content of each speaker. 
To further improve the SOT approach, several studies have proposed to use the auxiliary information of all the speakers that may appear in the speech signal, which is known as speaker-attribute ASR (SA-ASR)~\cite{kanda2021investigation,chang2021hypothesis,kanda2021minimum}.

Despite the tremendous progress in overlapping ASR, it is still challenging to achieve reasonable performance, particularly for weak speech. 
Providing an attention bias to indicate the target speech for recognition presents a more practical solution and aligns with the requirements of most real-world applications.
Speaker identity is a widely used bias~\cite{wang2019voicefilter,vzmolikova2019speakerbeam}, and the popular data augmentation approach actually signifies the model to decode the utterance with higher energy, assuming that the interference is low-energy babble noise. 
This research follows this research line and demonstrates that keywords, or specific spoken content of interest, can be used as an attention bias.

\section{Methods}
\label{sec:method}

\subsection{Overview of proposed methods}
\label{ssec:of}

\begin{figure}[htbp]
  \centering
     \includegraphics[width=0.92\linewidth, height=0.92\linewidth]{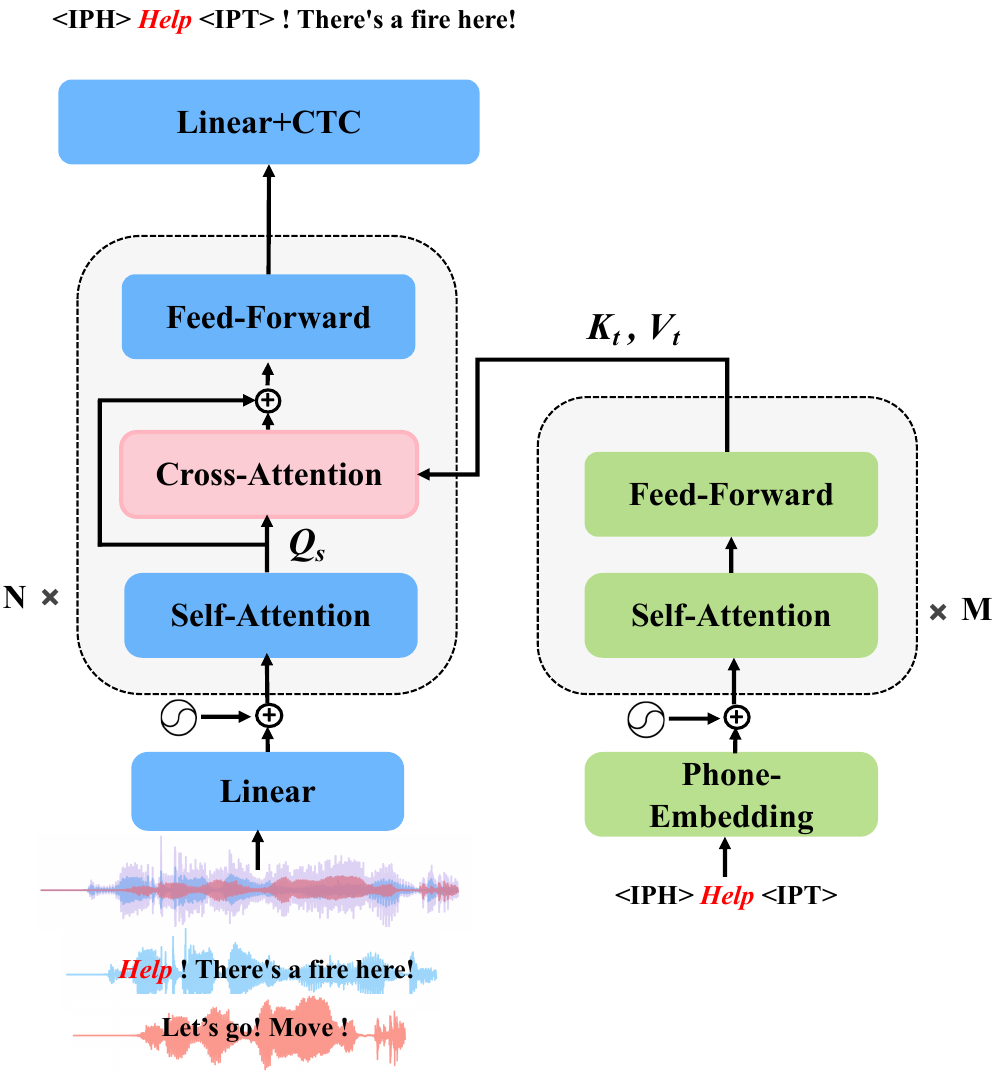}
    \caption{\textit{Overview of proposed methods}}
     \label{fig:overview}
\end{figure}

The entire architecture of the proposed TC-ASR model is shown in Fig.~\ref{fig:overview}.
The input speech, which could be a mixture of several utterances, is encoded by a speech encoder $f_{s}(\cdot)$, and the focused keyword is encoded by a keyword encoder $f_{k}(\cdot)$. 
Both encoders are implemented using Transformer modules, with integration achieved through cross-attention.
The output of the model is supposed to be the transcription of the utterance that contains the input keyword. 

Two crucial designs of the model are: (1) the attention mechanism, and (2) information pivoting. 
The former enables the selection of target speech from either concatenated or mixed speech, and the latter strengthens the attention to the keyword. 
More details will be provided in the following sections.

\subsection{Self-attention \& Cross-attention}
\label{ssec:ca}

The self-attention (SA) mechanism~\cite{vaswani2017attention} has been a crucial technique for encoding sequential data, including speech and text, mostly attributed to its capability to represent long-span correlations. 
In our TC-ASR model, the speech and keyword inputs are encoded by two Transformers, with SA as the primary building block.
The idea is that the long-span correlation will help extract \emph{coherent} speech component, in terms of both temporal continuity and timbre consistency. 

A key problem, however, is that Transformer cannot be directly adopted to encode overlapping speech. 
This is because the temporal correlation existing in a single speech is corrupted in overlapping speech. 
To solve the problem, we incorporate a cross-attention (CA) module into the Transformer structure, as shown by the pink block in Fig.~\ref{fig:overview}.
Specifically, we define the intermediate speech embedding at the $i$-th block of the speech encoder as query $(Q_{s}^i)$, and treat the keyword embedding as key $(K_{t})$ and value $(V_{t})$. The CA output can be expressed as follows: 

\begin{equation}
  \label{eq:ca}
  \text{CA}(Q_{s}^i, K_{t}, V_{t})=softmax(\frac{Q_{s}^i K^{T}_{t}}{\sqrt{d_{k}}})V_{t}
\end{equation}

Essentially, the cross-attention computation forces the model to pay attention to the component that contains the given keyword within the overlapping speech. 
This enforcement is strengthened at each block of the speech encoder, and the result is that the target speech gradually becomes more prominent in the forward pass, while any interference is gradually eliminated.

\subsection{Information Pivot}
\label{ssec:ip}

When encoding the keyword, a head token \textbf{\textless IPH\textgreater} and a tail token \textbf{\textless IPT\textgreater} are appended to the beginning and the end of the phone sequence, respectively.
These same tokens are also inserted into the transcription of the target speech. 
We hypothesize that these special tokens act as an information pivot, aiding the model in localizing the exact position of the keyword within the speech signal.
Experimental results demonstrated that the pivot tokens are crucial and their absence leads to significant performance degradation.

\section{Experiments}
\label{sec:exp}
\subsection{Data Preparation}
 
Experiments were performed using the Librispeech dataset~\cite{panayotov2015librispeech}.
The preparation of the training data and test data are presented below.

\textbf{Training Data:} First of all, utterances of all the training data in Librispeech were aligned to the corresponding word/phone sequences, using the MFA tools\footnote{https://montreal-forced-aligner.readthedocs.io}. 
To generate an overlapping speech training example, two utterances $U_1$ and $U_2$ were randomly sampled from the training set (officially the train\_960 set) and combined using weights $w_1$ and $w_2$, derived from a uniform distribution over the range [0.1,0.9], i.e., $w_1 \cdot U_1 + w_2 \cdot U_2$.
After that, a phrase consisting of 2-4 words was selected from the transcription of either $U_1$ or $U_2$ to serve as the keyword, with the corresponding full transcription used as the label for this training sample. More specifically, we formed both the keyword and the full transcription into phone sequences.

\textbf{Test Data:} In order to rigorously assess our method capability, we simulated a variety of test conditions with different SNRs.
Each overlapping test sample involves a target utterance that contains the focused keyword and an interfering utterance, both were sampled from the Librispeech test-clean dataset. 
Note that to raise the difficulty of the task, if the interference utterance was shorter than the target utterance, it was repeated until the two utterances were fully mixed. For each test sample, the keyword involves 3 words.

\subsection{Model Configuration}

\par
\textbf{Speech Encoder:} The speech signal is first converted to 40-dimensional Fbank features, which are input to the speech encoder. 
The first layer of the speech encoder is linear and transforms the speech Fbank features into 256-dimensional hidden features. 
A positional embedding is then added to the hidden features and the position-augmented features are forwarded to eight transformer blocks. 
Each transformer block incorporates a 4-head self-attention module, a 1-head cross-attention module, and a non-linear module that comprises a linear layer, layer normalization, and ReLU.

\textbf{Keyword Encoder:} The Keyword encoder initially embeds the phone sequence of the keyword into a series of 256-dimensional features. 
These features are then augmented by positional embedding and forwarded to four transformer blocks.
The structure of the transformer block in the keyword encoder is almost identical to that of the speech transformer block, with the exception that it lacks a cross-attention module.

\subsection{Implementation Details}
The input feature was a 40-dimensional Fbank with a window size of 25ms and a frame shift of 10ms. 
Before being fed into the neural network, the Fbank was spliced with a [-2,-1,0,1,2] context, rendering the input dimension of $f_{s}(\cdot)$ as 200. 
In practice, a 3-frame sub-sampling was also incorporated. 
For training, we utilized the CTC loss as our loss function and employed the Adam optimizer with a learning rate of 1e-3. 
All the models were trained over 80 epochs with a batch size of 32. A warm-up period was conducted over the initial 10 epochs. 
During testing, the final model is determined by taking the average of the checkpoints from the last ten epochs.

\section{Results}
\label{sec:res}

This section details the experimental results. 
The Clean model, trained exclusively on clean data, serves as one of our baseline models.
Additionally, we introduced two models trained with data augmentation:
DA-Strong was trained with samples comprising strong target speech and weak interference, while DA-Weak was trained with samples comprising weak target speech and strong interface. 
These two models use relative energy as the attention bias and are expected to perform well if the training and test conditions match. 

For the proposed TC-ASR approach, we trained three models: TC-Strong and TC-Weak, which were trained in a similar manner to DA-Strong and DA-Weak, respectively.
TC-Full means the model was trained using both strong and weak interference. 

The speech encoder was the same for all the DA and TC models (2.7M parameters), while the TC models involved a text encoder (1.4M parameters).

\subsection{Main results}

The main results with the clean baseline, the DA and TC models are presented in Table~\ref{tab:snrtest}, where the Signal-to-Noise Ratio (SNR) 
of the overlapping speech varies from -3dB to 3dB (target speech as the signal and interference speech as the noise). 
The results on clean speech are also presented. The phone sequence of a three-word keyword served as the keyword condition.
Note that the target and the interference utterances were from different speakers in this experiment.

\begin{table}[h!]
    \caption{PER(\%) results with different models on various SNR conditions.}
    \label{tab:snrtest}
    \centering
    \begin{tabular}{lcccc}
    \hline
    \toprule
      \multirow{1}{*}{Model}   &\multicolumn{1}{c}{Clean}   & \multicolumn{1}{c}{-3dB}   & \multicolumn{1}{c}{0dB}   & \multicolumn{1}{c}{3dB} \\
      \cmidrule(r){1-1}         \cmidrule(r){2-2}             \cmidrule(r){3-3}             \cmidrule(r){4-4}           \cmidrule(r){5-5}
          Clean                & \textbf{7.64}              &   95.73                     &  92.30                    &  87.49  \\
          DA-Strong            & 16.29                      &   117.00                    &  71.11                    &  22.70  \\
          DA-Weak              & 76.05                      &   27.62                     &  72.73                    &  116.29 \\
          TC-Strong            & 12.78                      &   73.81                     &  38.38                    &  \textbf{17.35}  \\
          TC-Weak              & 49.41                      &   \textbf{21.24}            &  35.73                    &  75.93  \\
          TC-Full              & 19.04                      &   26.06                     &  \textbf{23.18}           &  20.93 \\
    \bottomrule
    \hline
    \end{tabular}
  \end{table}


The first observation is that the baseline model achieved a Phone Error Rate (PER) of 7.64\%, which is reasonable considering the size of the model  (2.7M parameters).
All the DA and TC models perform worse than the clean model on clean speech, which is not surprising as no clean utterances were used in model training. 
However, once the target utterances are corrupted by the interference speech, the clean model simply fails. 

By focusing on the DA models, we find that when training and test conditions match, the performance of DA models significantly surpasses that of the clean baseline.
For example, DA-Strong achieves a PER of 22.70\% in the 3dB test, and DA-Weak achieves a PER of 27.62\% in the -3dB test, clearly superior to the clean model. 
However, if there is a mismatch between the training and test conditions, the performance of the DA models significantly deteriorates.
These results demonstrate that energy is a reasonable attention bias, and the model can learn to know which utterance to recognize based on the relative energy. If the bias is incorrect during the test, the model fails completely.

For the TC models, one can find that in all the test cases they perform better than the DA counterparts, i.e., TC-Strong works better than DA-Strong and TC-Weak works better than DA-Weak. This trend is more clear when the training and test conditions are mismatched. 
These observations clearly demonstrate that the keyword provides a highly effective attention bias that can guide the model to choose the correct utterance to perform recognition. 
For this reason, some target utterances can be successfully extracted by looking at the keyword even if the energy bias is wrong, which is the case in the training-test mismatch conditions. 
When trained with the full data, TC-Full yielded reasonable results in all the test cases. Particularly in the 0dB test set, the energy bias is not valid anymore and the only bias is the keyword. In this case, TC-Full yields a PER of 23.18\%, significantly outperforming other models. 

In summary, both energy and keywords can offer reasonable attention bias, with which the model can identify the target utterance from overlapping speech. 
If we need to recognize target utterances whose energy could be either strong or weak, the keyword is a very effective bias. 

\subsection{Analysis study}

\begin{figure}[htbp]
  \centering
     \includegraphics[width=0.96\linewidth]{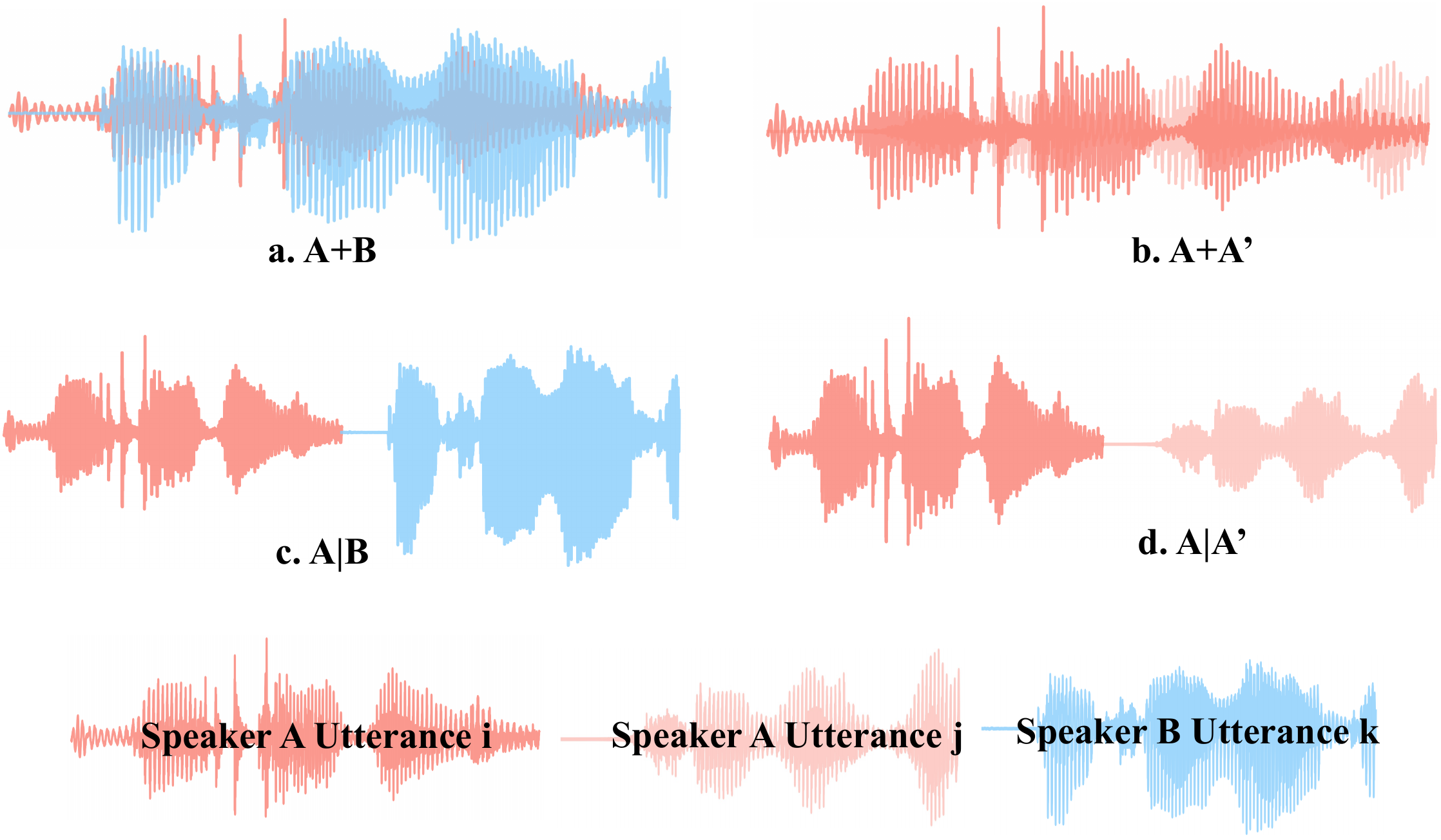}
    \caption{\textit{Illustration of mixing and concatenation. \textbf{A} and \textbf{B} denote
    utterances from two speakers, while \textbf{A} and \textbf{A'} represent two utterances 
    from the same speaker. `+' means speech mixing and `$|$' means speech concatenation.}}
     \label{fig:mix}
\vspace{-3.0mm}
\end{figure}

We have demonstrated that the proposed TC model can extract target utterances by `glancing' a keyword. 
The question then arises: how the model can achieve this?
There are two possibilities: (1) Temporal continuity, where the model detects words surrounding the keyword to form a continuous speech and word sequence; (2) Speaker identity, where the model detects words spoken by the same person who speaks the keyword.
To test these two hypotheses, we designed a set of mixing and concatenating 
experiments, as depicted in Figure~\ref{fig:mix}. 
It should be noted that in all the experiments, the transcription of utterance A is regarded as the desired output. 

\begin{table}[htb!]
    \caption{PER(\%) results of TC-Full with mixing and concatenation. }
    \label{tab:mix_concat}
    \centering
    \scalebox{0.90}{
    \begin{tabular}{rcccc}
    \hline
    \toprule
      \multirow{1}{*}{SNR}   &\multicolumn{1}{c}{A+B}   & \multicolumn{1}{c}{A+A'}   & \multicolumn{1}{c}{A$|$B}   & \multicolumn{1}{c}{A$|$A'} \\
      \cmidrule(r){1-1}         \cmidrule(r){2-2}             \cmidrule(r){3-3}             \cmidrule(r){4-4}           \cmidrule(r){5-5}
            -3dB            &  26.06                    & 44.71                      &  14.92                      &  74.96   \\
             0dB             &  23.18                   & 44.77                      &  14.85                      & 71.07  \\
             3dB             &  20.93                   & 42.12                      &  15.03                      & 58.01 \\
    \bottomrule
    \hline
    \end{tabular}
    } 
  \end{table}

The results of the TC-Full model on these test conditions are shown in Table~\ref{tab:mix_concat}.
Here, the SNR only refers to the relative energy between the target and interference utterances, irrespective of whether they overlap.

Note that the A+B condition is just the condition tested in the main experiment. 
A critical observation is that across all SNR conditions, mixing an additional utterance from the same target person (A+A') results in significantly poorer performance than when mixing an utterance from a non-target person.
This suggests that the model depends on speaker identity to extract the target utterance from the mixed signal, thereby strongly supporting the speaker identity hypothesis over the temporal continuity hypothesis.

Furthermore, concatenating an utterance from a non-target speaker (A$|$B) yields a PER of 15.03\%, even better than the results on clean utterances. 
This implies that nearly all the signals from B are ignored by the decoder. 
In contrast, concatenating an utterance from the same target speaker leads to rather poor performance, and most of them are insertion errors.
This indicates that the decoder identified both A and A' as target utterances. 
Once again, this is consistent with the speaker identity hypothesis. 
This experiment also demonstrated that the TC-Full model can identify the target speech from not only overlapping utterances but also concatenated utterances if the interference utterance is from different speakers. 

\begin{table}[h!]
  \caption{PER(\%) of TC-Full with different keyword lengths.}
  \label{tab:wordlen}
  \centering
 \scalebox{0.90}{
  \begin{tabular}{lccc}
  \toprule
    \multirow{1}{*}{Model}      & \multicolumn{1}{c}{-3dB}   & \multicolumn{1}{c}{0dB}   & \multicolumn{1}{c}{3dB} \\
    \cmidrule(r){1-1}                     \cmidrule(r){2-2}             \cmidrule(r){3-3}           \cmidrule(r){4-4}
        1-word                                &   41.26                     &  39.24                    &  37.79  \\
        2-word                                &   29.56                     &  26.70                    &  24.20  \\
        3-word                                &   26.06                     &  23.18                    &  20.93 \\
  \bottomrule
  \hline
  \end{tabular}
 } 
\end{table}

Table~\ref{tab:wordlen} compares the impact of keyword length on the performance of the TC-Full model. 
It should be noted that the 3-word case is the default setting and was used in previous experiments. 
The results show that performance is degraded when the keyword is short. 
This is also consistent with the speaker identity assumption: with a short keyword, identifying the position of the keyword segment is 
hard and the speaker identity inferred from a short speech segment is unreliable.

\begin{table}[h!]
  \caption{PER(\%) with/without pivot tokens.}
  \label{tab:pivot}
  \centering
  \scalebox{0.95}{
  \begin{tabular}{lccc}
    \toprule
    \multirow{1}{*}{Model}      & \multicolumn{1}{c}{-3dB}   & \multicolumn{1}{c}{0dB}   & \multicolumn{1}{c}{3dB} \\
    \cmidrule(r){1-1}            \cmidrule(r){2-2}             \cmidrule(r){3-3}           \cmidrule(r){4-4} 
        TC-Full                 &   26.06                     &  23.18                    &  20.93 \\
        TC-Full - No Pivot      &   39.97                     &  37.12                    &  34.69 \\
  \bottomrule
  \hline
  \end{tabular}
  } 
\end{table}

Finally, we evaluate the importance of the pivot tokens for the TC-Full model. 
The results are shown in Table~\ref{tab:pivot}, where each keyword contains 3 words. 
It can be seen that the pivot tokens are crucial for the TC model. 

\section{Conclusion \& Furthur Work}

In this study, we investigate the possibility of using a keyword as an attention bias to notify a neural model to identify and recognize target speech from multi-talker speech, which may involve overlapping and concatenation of speech from multiple speakers.
A simple model that comprises Transforms and the cross-attention mechanism was proposed.
The results are highly promising: Our model can identify and recognize speech containing the keyword by looking at the keyword, regardless of whether the interference speech overlaps or concatenates with the target speech.

Further analysis shows that the underlying mechanism of the target-content ASR is that the model essentially learns the characteristics of the speaker who speaks the keyword by identifying the position of the keyword extracting its speech segment, and then recognizing the entire utterance of that speaker using the speaker identity.
The second part of the story has been demonstrated by target-speaker ASR, where speaker identity is used as the attention bias, but it is seldom known that a keyword can be used as an attention bias directly. 
In some circumstances, using a keyword as the bias is clearly more suitable. 

It should be noted that the current study is still preliminary, and experiments with other datasets need to be conducted to confirm the discovery.
Additionally, we have not yet experimented with input speech that does not contain any keywords.
We also need to study if the target speech can be recovered from mixed or concatenated speech -- if the speaker identity assumption is correct, this should be feasible. 
Finally, joint training for both the ASR task and keyword spotting task also deserves further investigation, as the ASR output may provide useful information for keyword spotting to reduce false alarms. 

\vfill

\bibliographystyle{IEEEbib}
\bibliography{strings,refs}

\begin{thebibliography}{10}

\bibitem{xiong2016achieving}
Wayne Xiong, Jasha Droppo, Xuedong Huang, Frank Seide, Mike Seltzer, Andreas
  Stolcke, Dong Yu, and Geoffrey Zweig,
\newblock ``Achieving human parity in conversational speech recognition,''
\newblock {\em arXiv preprint arXiv:1610.05256}, 2016.

\bibitem{park2022review}
Tae~Jin Park, Naoyuki Kanda, Dimitrios Dimitriadis, Kyu~J Han, Shinji Watanabe,
  and Shrikanth Narayanan,
\newblock ``A review of speaker diarization: Recent advances with deep
  learning,''
\newblock {\em Computer Speech \& Language}, vol. 72, pp. 101317, 2022.

\bibitem{chen2017progressive}
Zhehuai Chen, Jasha Droppo, Jinyu Li, and Wayne Xiong,
\newblock ``Progressive joint modeling in unsupervised single-channel
  overlapped speech recognition,''
\newblock {\em IEEE/ACM Transactions on Audio, Speech, and Language
  Processing}, vol. 26, no. 1, pp. 184--196, 2017.

\bibitem{kawata2020neural}
Natasha Yuriko~Santos Kawata, Teruo Hashimoto, and Ryuta Kawashima,
\newblock ``Neural mechanisms underlying concurrent listening of simultaneous
  speech,''
\newblock {\em Brain research}, vol. 1738, pp. 146821, 2020.

\bibitem{wang2019voicefilter}
Quan Wang, Hannah Muckenhirn, Kevin Wilson, Prashant Sridhar, Zelin Wu, John~R
  Hershey, Rif~A Saurous, Ron~J Weiss, Ye~Jia, and Ignacio~Lopez Moreno,
\newblock ``Voicefilter: Targeted voice separation by speaker-conditioned
  spectrogram masking,''
\newblock {\em Proc. Interspeech 2019}, pp. 2728--2732, 2019.

\bibitem{vzmolikova2019speakerbeam}
Kate{\v{r}}ina {\v{Z}}mol{\'\i}kov{\'a}, Marc Delcroix, Keisuke Kinoshita,
  Tsubasa Ochiai, Tomohiro Nakatani, Luk{\'a}{\v{s}} Burget, and Jan
  {\v{C}}ernock{\`y},
\newblock ``Speakerbeam: Speaker aware neural network for target speaker
  extraction in speech mixtures,''
\newblock {\em IEEE Journal of Selected Topics in Signal Processing}, vol. 13,
  no. 4, pp. 800--814, 2019.

\bibitem{moriya2022streaming}
Takafumi Moriya, Hiroshi Sato, Tsubasa Ochiai, Marc Delcroix, and Takahiro
  Shinozaki,
\newblock ``Streaming target-speaker asr with neural transducer,''
\newblock in {\em Proc. of INTERSPEECH}, 2022, pp. 2673--2677.

\bibitem{zhang2023conformer}
Yang Zhang, Krishna~C Puvvada, Vitaly Lavrukhin, and Boris Ginsburg,
\newblock ``Conformer-based target-speaker automatic speech recognition for
  single-channel audio,''
\newblock in {\em ICASSP 2023-2023 IEEE International Conference on Acoustics,
  Speech and Signal Processing (ICASSP)}. IEEE, 2023, pp. 1--5.

\bibitem{wu2021investigation}
Jian Wu, Zhuo Chen, Sanyuan Chen, Yu~Wu, Takuya Yoshioka, Naoyuki Kanda, Shujie
  Liu, and Jinyu Li,
\newblock ``Investigation of practical aspects of single channel speech
  separation for asr,''
\newblock {\em arXiv preprint arXiv:2107.01922}, 2021.

\bibitem{yu2017recognizing}
Dong Yu, Xuankai Chang, and Yanmin Qian,
\newblock ``Recognizing multi-talker speech with permutation invariant
  training,''
\newblock {\em Interspeech 2017}, vol. 2017, pp. 2456, 2017.

\bibitem{settle2018end}
Shane Settle, Jonathan Le~Roux, Takaaki Hori, Shinji Watanabe, and John~R
  Hershey,
\newblock ``End-to-end multi-speaker speech recognition,''
\newblock in {\em 2018 IEEE international conference on acoustics, speech and
  signal processing (ICASSP)}. IEEE, 2018, pp. 4819--4823.

\bibitem{7122291}
Chao Weng, Dong Yu, Michael~L. Seltzer, and Jasha Droppo,
\newblock ``Deep neural networks for single-channel multi-talker speech
  recognition,''
\newblock {\em IEEE/ACM Transactions on Audio, Speech, and Language
  Processing}, vol. 23, no. 10, pp. 1670--1679, 2015.

\bibitem{lu2021streaming}
Liang Lu, Naoyuki Kanda, Jinyu Li, and Yifan Gong,
\newblock ``Streaming end-to-end multi-talker speech recognition,''
\newblock {\em IEEE Signal Processing Letters}, vol. 28, pp. 803--807, 2021.

\bibitem{yu2017permutation}
Dong Yu, Morten Kolb{\ae}k, Zheng-Hua Tan, and Jesper Jensen,
\newblock ``Permutation invariant training of deep models for
  speaker-independent multi-talker speech separation,''
\newblock in {\em 2017 IEEE International Conference on Acoustics, Speech and
  Signal Processing (ICASSP)}. IEEE, 2017, pp. 241--245.

\bibitem{kolbaek2017multitalker}
Morten Kolb{\ae}k, Dong Yu, Zheng-Hua Tan, and Jesper Jensen,
\newblock ``Multitalker speech separation with utterance-level permutation
  invariant training of deep recurrent neural networks,''
\newblock {\em IEEE/ACM Transactions on Audio, Speech, and Language
  Processing}, vol. 25, no. 10, pp. 1901--1913, 2017.

\bibitem{qian2018single}
Yanmin Qian, Xuankai Chang, and Dong Yu,
\newblock ``Single-channel multi-talker speech recognition with permutation
  invariant training,''
\newblock {\em Speech Communication}, vol. 104, pp. 1--11, 2018.

\bibitem{meng2023sidecar}
Lingwei Meng, Jiawen Kang, Mingyu Cui, Yuejiao Wang, Xixin Wu, and Helen Meng,
\newblock ``A sidecar separator can convert a single-talker speech recognition
  system to a multi-talker one,''
\newblock in {\em ICASSP 2023-2023 IEEE International Conference on Acoustics,
  Speech and Signal Processing (ICASSP)}. IEEE, 2023, pp. 1--5.

\bibitem{kanda2020serialized}
Naoyuki Kanda, Yashesh Gaur, Xiaofei Wang, Zhong Meng, and Takuya Yoshioka,
\newblock ``Serialized output training for end-to-end overlapped speech
  recognition,''
\newblock {\em Proc. Interspeech 2020}, pp. 2797--2801, 2020.

\bibitem{kanda2021investigation}
Naoyuki Kanda, Xuankai Chang, Yashesh Gaur, Xiaofei Wang, Zhong Meng, Zhuo
  Chen, and Takuya Yoshioka,
\newblock ``Investigation of end-to-end speaker-attributed asr for continuous
  multi-talker recordings,''
\newblock in {\em 2021 IEEE Spoken Language Technology Workshop (SLT)}. IEEE,
  2021, pp. 809--816.

\bibitem{chang2021hypothesis}
Xuankai Chang, Naoyuki Kanda, Yashesh Gaur, Xiaofei Wang, Zhong Meng, and
  Takuya Yoshioka,
\newblock ``Hypothesis stitcher for end-to-end speaker-attributed asr on
  long-form multi-talker recordings,''
\newblock in {\em ICASSP 2021-2021 IEEE International Conference on Acoustics,
  Speech and Signal Processing (ICASSP)}. IEEE, 2021, pp. 6763--6767.

\bibitem{kanda2021minimum}
Naoyuki Kanda, Zhong Meng, Liang Lu, Yashesh Gaur, Xiaofei Wang, Zhuo Chen, and
  Takuya Yoshioka,
\newblock ``Minimum bayes risk training for end-to-end speaker-attributed
  asr,''
\newblock in {\em ICASSP 2021-2021 IEEE International Conference on Acoustics,
  Speech and Signal Processing (ICASSP)}. IEEE, 2021, pp. 6503--6507.

\bibitem{vaswani2017attention}
Ashish Vaswani, Noam Shazeer, Niki Parmar, Jakob Uszkoreit, Llion Jones,
  Aidan~N Gomez, {\L}ukasz Kaiser, and Illia Polosukhin,
\newblock ``Attention is all you need,''
\newblock {\em Advances in neural information processing systems}, vol. 30,
  2017.

\bibitem{panayotov2015librispeech}
Vassil Panayotov, Guoguo Chen, Daniel Povey, and Sanjeev Khudanpur,
\newblock ``Librispeech: an asr corpus based on public domain audio books,''
\newblock in {\em 2015 IEEE international conference on acoustics, speech and
  signal processing (ICASSP)}. IEEE, 2015, pp. 5206--5210.

\end{thebibliography}

\end{document}